\documentclass[letterpaper, 10 pt, conference]{ieeeconf}  

\IEEEoverridecommandlockouts                              

\usepackage{graphics} 
\usepackage{epsfig} 
\usepackage{mathptmx} 
\usepackage{times} 
\usepackage{amsmath} 
\usepackage{amssymb}  
\usepackage{cite}
\usepackage{caption}
\usepackage{subcaption}
\usepackage{algorithm}
\usepackage[noend]{algpseudocode}

\title{\LARGE \bf
Optimization Based Collision Avoidance for Multi-Agent Dynamical Systems in Goal Reaching Task
}

\author{Adarsh Patnaik$^{1}$ and Ashish Ranjan Hota$^{2}$
\thanks{$^{1}$Department of Mechanical Engineering, Indian Institute of Technology Kharagpur}
\thanks{$^{2}$Department of Electrical Engineering, Indian Institute of Technology Kharagpur}
}

\begin{document}

\maketitle
\thispagestyle{empty}
\pagestyle{empty}

\begin{abstract}

This work presents a distributed MPC-based approach to solving the problem of multi-agent point-to-point transition with optimization-based collision avoidance. The problem is formulated motivated by the work on collision avoidance for multi-agent systems and dynamic obstacles. With modifications to the formulation, the problem is converted into a distributed problem with a separable objective and coupled constraints. The problem is divided into local sub-problems
and solved using Alternating Directions Method of Multipliers (ADMM) applied on an augmented local lagrangian objective. This work aims to understand the multi-agent point-to-point transition problem as an extension of optimization-based collision avoidance and analyze the aspects of computational times, reliability, and optimality of the solution obtained.
\end{abstract}

\section{INTRODUCTION} \label{introduction}
Trajectory optimization for robotics is a very common problem which has seen several advancements over time. From providing simple hand designed trajectories, to graph based search methods and random sampling methods, and finally the use of high level analytical and numerical optimization methods for optimizing state space trajectories. With the availability of computational resources, Model Predictive Control (MPC) has been recently used for online trajectory optimization. With the use of MPC, the extensive knowledge of the dynamics of various robots can now be used to perform model based optimization. This expands to a variety of systems with linear, non-linear and hybrid dynamics where MPC can be used to optimize trajectories in real time with robust constraint satisfaction.

Recently, MPC has also been used for multi-agent systems with the introduction of distributed MPC approach. One of the key challenges associated with trajectory optimization for multi-agent systems is efficient cooperation among the agents to maximize a global objective while satisfying local constraints like collision avoidance with other agents. This problem is hard to solve due to the non-convex objectives and constraints associated with it and due to scalability issues with higher number of agents. We consider the problem of a swarm of agents modelled as point masses with the objective of point to point transitions for each of the agent. This problem has a local objective for each of the agents while it also needs to satisfy global constraints like collision avoidance dependant on the dynamics of other agents. Several formulations and optimization methods have been discussed for similar problems. We model the problem with linearly separable objective and global coupled collision avoidance constraints and use distributed MPC for solving the global problem in a distributed fashion.

Distributed MPC based approaches aim to use consensus based distributed optimization algorithms that can help to divide the large complex problem in smaller sub-problems which can be handled by each agent separately. Optimization methods like Alternating Direction Method of Multipliers (ADMM) have been used to solve the distributed MPC problem and have proved to be computationally efficient for real time operations. In this work, we solve the multi-agent point to point transition problem using an ADMM based distributed MPC algorithm. The key contributions of this work is are (1) An extension of the optimization based collision avoidance constraints to multi-agent systems, (2) Formulation of a distributed optimization problem for multi-agent point to point transitions, and (3) an ADMM based distributed MPC approach to solve the given optimization problem.

We first discuss the problem formulation in the section \ref{problem_form}. Within the problem formulation, we first look at the system dynamics considered for our problem. Next we discuss the single agent objective and the obstacle avoidance constraints for each of the agent. Using this information we formulate the global optimization problem with linearly separable objective and coupled constraints. Next we present the formulation of the global problem in a distributed form and the distributed optimization method used to solve the problem. Section \ref{exp_and_res} discusses the details of the practical implementation and experiments conducted to show the effectiveness of the method and the analysis of the results in simulation. We finally discuss the conclusions and the future work in Section \ref{conc_and_future}.

\section{RELATED WORK} \label{related_work}
Trajectory optimization for multi-agent systems is a growing field of research attracting a lot of attention due to the significant applications of multi-agent systems of robots. The work related to the field spans across different problem statements and applications. Considerable work has been done on trajectory optimization for aerial robotic swarms in different scenarios. Works like \cite{Sung2018DistributedSA, Bayram2017TrackingWW} focus on formation control and multi-agent path following tasks which are becoming increasingly popular. Several work like \cite{Tang2015MixedIQ, Foehn2017FastTO} consider the problem of transporting heavy suspended payloads using a swarm of aerial robots which combines the use of geometric control and distributed optimal control. Various exploration and target searching applications also have considerable works like \cite{Bandyopadhyay2018DistributedBF, Bayram2017TrackingWW}. A variety of distributed methods \cite{Parys2017DistributedMP, Park2019ADA} based on Alternating Direction Method of Multipliers (ADMM) and mixed integer programs have also come up on how to optimize such optimization problems.

Optimization based collision avoidance strategies require to incorporate the condition into a trajectory optimization problem like MPC. Several work \cite{Khatib1985RealtimeOA,Zucker2013CHOMPCH, Li2016BiRRTOptAC} try to get rid of the non-convexity by modelling obstacles as repulsive potentials like gaussian and augmenting it to the objective function to penalize getting close to obstacles but such methods can lead to local optima and infeasible solutions a large number of times. Recent works model the obstacle collision condition as constraints \cite{Schulman2014MotionPW}. This helps to bring formal guarantees for collision avoidance at the cost of difficult optimization. The obstacles are generally modelled as convex polyhedrons or ellipsoids in order to mathematically formulate the constraints. Optimization methods like mixed integer programming \cite{Blackmore2006OptimalMP} and non linear programming \cite{Rosolia2017AutonomousVC, Naegeli2017RealTimeMP} are used to solve the optimization problems with collision avoidance constraints.

Several works focus on solving the multi-agent problem in a distributed fashion. The collision constraints between agents can be handled easily in such methods \cite{Bhattacharya2010DistributedOW, Rezaee2014ADC} and the computational load reduces. Works like \cite{Berg2011ReciprocalCA, AlonsoMora2010OptimalRC} introduce guaranteed collision avoidance for different type of dynamic systems at the cost of conservativeness. Distributed MPC based methods have also been used in multi-agent trajectory optimization \cite{Parys2017DistributedMP, Sayyaadi2017DecentralizedPT, Mellinger2011MinimumST} for a range of problems. \cite{Luis2019TrajectoryGF, Luis2020OnlineTG} discuss the use of distributed MPC for the point to point transition problem along with obstacle avoidance.

\section{PROBLEM FORMULATION}\label{problem_form}
In this section we discuss the various aspects of the problem in hand and the adopted solution method along with certain caveats supplementing the current solution. We first have a look at the system and environment descriptions followed by the optimization objectives for each of the agent. We also look at the various constraint handling methods for the obstacle avoidance criterion. Further, we formulate the problem in a multi-agent scenario with global communication. Then we discuss the distributed formulation and the Alternating Direction Method of Multipliers (ADMM) used to solve the given distributed problem. Next, we discuss some implementation details required to efficiently solve the problem.

\subsection{System Dynamics}
Within this work, we take into consideration two different set of systems for validating our solution methods. We consider a simple differential drive robot satisfying non holonomic constraints and navigating in a 2D environment with area constraint. Next, we consider a quadrotor aerial drone navigating in a 3D environment with volume constraint. Various complex dynamics model can be used to analyze these systems but for the simplicity of the problem, we model these systems using a double integrator in 2D and 3D respectively. We assume the availability of a position controller in both cases such that providing either accelerations as inputs or position points as the trajectory can work on the real robot.
\begin{equation}
    \Dot{s} = A.s + B.u
\end{equation}
where for $s = [x, y, v_{x}, v_{y}]$ and $u = [a_{x}, a_{y}]$
\begin{subequations}
    \begin{align}
        &A = \begin{bmatrix}0 & 0 & 1 & 0 \\ 0 & 0 & 0 & 1 \\0 & 0 & 0 & 0 \\0 & 0 & 0 & 0 \end{bmatrix} \\
        &B = \begin{bmatrix}0 & 0 \\0 & 0 \\1 & 0 \\0 & 1\end{bmatrix}
    \end{align}
\end{subequations}

and for $s = [x, y, z, v_{x}, v_{y}, v_{z}]$ and $u = [a_{x}, a_{y}, a_{z}]$
\begin{subequations}
    \begin{align}
        &A = \begin{bmatrix}0 & 0 & 0 & 1 & 0 & 0 \\0 & 0 & 0 & 0 & 1 & 0 \\0 & 0 & 0 & 0 & 0 & 1 \\0 & 0 & 0 & 0 & 0 & 0 \\ 0 & 0 & 0 & 0 & 0 & 0 \\ 0 & 0 & 0 & 0 & 0 & 0 \end{bmatrix} \\
        &B = \begin{bmatrix}0 & 0 & 0 \\0 & 0 & 0\\0 & 0 & 0\\1 & 0 & 0 \\0 & 1 & 0\\0 & 0 & 1\end{bmatrix}
    \end{align}
\end{subequations}

We discretize the double integrator dynamics for our work using the first order Euler discretization in order to get the following system dynamics equations.
\begin{subequations}
\begin{align}
    s_{k+1} = &A_{k}.s_{k} + B_{k}u_{k} \\
    A_{k} = &A.\Delta t + I_{n} \\
    B_{k} = &B.\Delta t
\end{align}
\end{subequations}

where $[A,B]$ are the respective continuous time matrices and $I_{n}$ is the order $n$ identity matrix for $n = \{2, 3\}$

\subsection{Single Agent Objective}\label{single_obj}
We consider the problem of point to point transition for a single agent. Starting from an initial state, the goal of the agent is to reach the specified goal state $s_{g}$ while avoiding other agents and any obstacles. We use a weighted quadratic cost term to formulate the objective. We also add a regularization term to limit the actuator values.
\begin{equation}\label{single_obj_eq}
\begin{split}
    &C(\textbf{s}, \textbf{u}) = \sum_{k=0}^{k=N-1} c(s(k),u(k)) + V(s(N)) \\
    &\textbf{s} = [s(0), s(1).....s(N)], \textbf{u} = [u(0), u(1).....u(N-1)]
\end{split}
\end{equation}
where 
\begin{subequations}
\begin{align}
    c(s,u) = &(s-s_{g})^{T}.Q.(s-s_{g}) + u^{T}.R.u \\
    V(s) = &(s-s_{g})^{T}.Q_{f}.(s-s_{g})
\end{align}
\end{subequations}
    
The value function $V(s)$ is initially considered to be the normal quadratic value function similar to the discrete time LQR problem. The use of the value function here is to prevent the solution from diverging away from the global objective and acts as a heuristic for the global optimization problem. Pre-computed heuristics for the value function can also be incorporated within the method to improve overall performance.

\subsection{Obstacle Avoidance Constraint}

\subsubsection{Obstacle Modelling}
In this work, we model the obstacles as convex compact sets $\mathbb{O}$. We represent the obstacle space as
\begin{equation}
    \mathbb{O} = \{ s\in \mathbb{R}^{n}: G.s \leq g \}
\end{equation}
where $G \in \mathbb{R}^{lxn}$, $g \in \mathbb{R}^{l}$. Any polyhedral obstacle can be represented in the form shown above. As discussed in \cite{Zhang2017OptimizationBasedCA}, this formulation is generalized for a convex pointed cone with non empty interior. For our work, we consider a second order cone and hence the general formulation converts to the common $\leq$ term.

For obstacle avoidance we need to ensure the following non differentiable constraint to satisfy
\begin{equation}
    s \cap \mathbb{O} = \phi
\end{equation}
where $s$ is the position of the agent. This constraint is reformulated as given in \cite{Zhang2017OptimizationBasedCA} to preserve continuity and differentiability.

\subsubsection{Signed Distance Method}

\begin{figure}
    \centering
    \begin{subfigure}[b]{0.23 \textwidth}
    \centering
    \includegraphics[width=\textwidth]{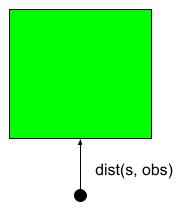}
    \caption{Estimation of $dist(s,\mathbb{O})$}
    \label{dist}
    \end{subfigure}
    \begin{subfigure}[b]{0.23 \textwidth}
    \centering
    \includegraphics[width=\textwidth]{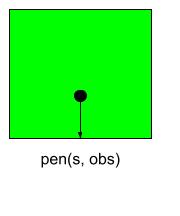}
    \caption{Estimation of $pen(s,\mathbb{O})$}
    \label{pen}
    \end{subfigure}
    \caption{Schematic of the $dist(.,.)$ and $pen(.,.)$ values}
    \label{sign_dist}
\end{figure}

The signed distance method is a common method for obstacle avoidance. The sign of the signed distance function provides the indication of whether we have a collision with an obstacle. The signed distance function is defined as follows.
\begin{equation}
    sdf(s,\mathbb{O}) = dist(s,\mathbb{O}) - pen(s,\mathbb{O})
\end{equation}
where
\begin{subequations}
\begin{align}
    dist(s, \mathbb{O}) :=& \min_{t} \{ ||t|| : (s+t) \cap \mathbb{O} \neq \phi \} \\
    pen(s, \mathbb{O}) :=& \min_{t} \{ ||t|| : (s+t) \cap \mathbb{O} = \phi \}
\end{align}
\end{subequations}
Here $dist(s,\mathbb{O})$ gives us the minimum distance from the polyhedron when the point is outside and $pen(s, \mathbb{O})$ gives the minimum distance from the polyhedron when the point is inside as shown in Fig. \ref{sign_dist}.
For preventing collision with the obstacle, the $sdf(s, \mathbb{O}) \geq 0$. But as the $sdf(.,.)$ is the result of an optimization problem, it leads to a bilevel optimization problem with the result of one constraining the other which is difficult to solve. So it needs to be reformulated before it can be included in the optimization problem.

\subsubsection{Collision Avoidance Reformulation}\label{section:collision_avoidance_formulation}
As discussed in \cite{Zhang2017OptimizationBasedCA}, two different formulations for the collision avoidance constraint are provided by modelling the agent as point mass and the obstacle as discussed above. We discuss both the formulations ahead. The first formulation ensures that the distance between the obstacle and the agent is greater than a minimum threshold.
\begin{equation}\label{equation:standard_formulation}
\begin{split}
    &dist(s, \mathbb{O}) \geq d_{min} \\
    & \Longleftrightarrow \exists \lambda \geq 0 : (Gs-g)^{T}\lambda \geq d_{min}, ||G^{T}\lambda||_{2} \leq 1
\end{split}
\end{equation}
Here $\lambda$ acts as a certificate for collision avoidance. Using this the optimal control problem for a single agent can be written as

\begin{equation}
\begin{split}
    \min_{\textbf{s},\textbf{u},\lambda} & \sum_{k=0}^{k=N-1} c(s(k),u(k)) + V(s(N)) \\
    s.t. \hspace{4pt} & s(k+1) = A_{k}.s(k) + B_{k}.u(k) \\
    & (Gs(k)-g)^{T}.\lambda_{k} \geq 0 \\
    & ||G^{T}\lambda_{k}||_{2} \leq 1, \lambda_{k} \geq 0 \\
\end{split}
\end{equation}

    

The second formulation gives a less conservative approach to model the collision avoidance constraint. It is based on the logic of keeping the penetration depth of the agent inside the obstacle space less than a maximum threshold.

\begin{equation} \label{equation:min_penetration}
\begin{split}
    &pen(s, \mathbb{O}) \leq p_{max} \\
    & \Longleftrightarrow \exists \lambda \geq 0 : (g-Gs)^{T}\lambda \leq p_{max}, ||G^{T}\lambda||_{2} = 1
\end{split}
\end{equation}

This formulation allows the signed distance function to be negative as well and hence provides less conservative results but the constraint $||G^{T}\lambda||_{2} = 1$ adds non-convexity to the problem which leads to additional computational load and accurate initial guesses to be solved optimally. Using this minimum penetration reformulation, the optimal control problem can be written as,

\begin{equation}
\begin{split}
    \min_{\textbf{s},\textbf{u},\lambda,\alpha} &\sum_{k=0}^{k=N-1} c(s(k),u(k)) + \kappa.\alpha_{k} + V(s(N)) \\
    s.t. \hspace{4pt} & s(k+1) = A_{k}.s(k) + B_{k}.u(k)\\
    & (Gs(k)-g)^{T}.\lambda_{k} \geq -\alpha_{k}\\
    & ||G^{T}\lambda_{k}||_{2} = 1, \lambda_{k} \geq 0, \alpha_{k} \geq 0
\end{split}
\end{equation}

where $\alpha_{k}$ denotes the slack variable which must be very close to zero and should only become active when collision cannot be avoided and a minimum penetration trajectory is to be obtained. We use both the formulations to experiment between computational load and feasibility for different scenarios for our work.

\subsection{Distributed Formulation}

In this section we formulate the problem for the multi-agent system and discuss the objective and constraints associated with the global problem for the entire swarm.

\subsubsection{Global Objective}
The global objective is formulated in such a way that it is separable and can be solved in a distributed fashion. Taking the single agent objectives from section \ref{single_obj}, we use a simple summation of each of the single agent objectives to form the global objective.
\begin{equation}
    \textbf{C} = \sum_{m=0}^{m=M} C(\textbf{s}^{(m)}, \textbf{u}^{(m)})
\end{equation}
where $C(.,.)$ and $\textbf{s}^{(m)}, \textbf{u}^{(m)}$ are taken from Equation \ref{single_obj_eq} for the $m$th agent. We observe that this objective is separable and can be solved in a distributed fashion with each agent solving it's own local objective.

\subsubsection{Modelling Agents As Dynamic Obstacles}
\begin{figure}[!tbh]
    \centering
    \includegraphics[width= 0.45 \textwidth]{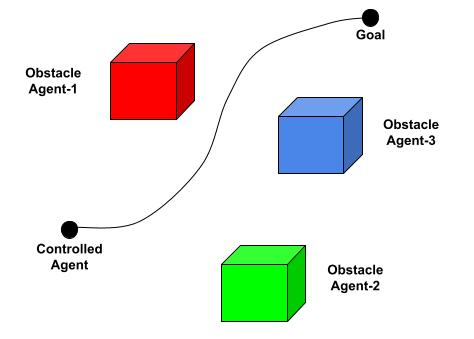}
    \caption{Schematic of Modelling Agents as Cubic Obstacles}
    \label{obstacle_avoidance_schematic}
\end{figure}

For a distributed algorithm for a multi-agent system, each agent solves a finite time optimal control problem independently of the other agents. Exploiting this nature of the method, we model each agent as a cube around the spatial position of the agent as shown in Fig \ref{obstacle_avoidance_schematic}. This cube gives the obstacle space as discussed before in the single agent case. This can be written mathematically as
\begin{equation}\label{cube}
    \Tilde{s}^{(m)} - \Delta \leq \Tilde{s} \leq \Tilde{s}^{(m)} + \Delta
\end{equation}
where $\Tilde{s}$ means the spatial position vector, $\Delta$ is the threshold distance, and $m \in \{0,1,...M \}$. For 2D system this can be also written as

\begin{equation}
    \begin{bmatrix}1 & 0\\ 0 & 1 \\ -1 & 0 \\ 0 & -1\end{bmatrix}(\Tilde{s}-\Tilde{s}^{(m)}) \leq \Delta
\end{equation}
and similar representation can be made in 3D as well.
Now using the reformulation technique discussed before, we can formulate the obstacle collision avoidance between agents. For each agent, every other agent behaves as an obstacle cube represented by the equation \ref{cube}. Using this we can formulate the obstacle collision avoidance for the $i$th agent as

\begin{subequations}
\begin{align}
    \Bigg(\begin{bmatrix}1 & 0\\ 0 & 1 \\ -1 & 0 \\ 0 & -1\end{bmatrix} (\Tilde{s}^{(i)}(k) - \Tilde{s}^{(j)}(k)) - \Delta \Bigg)^{T} \lambda_{k}^{j} &\geq -\alpha_{k}^{j} \\
    \Bigg|\Bigg|\begin{bmatrix}1 & 0\\ 0 & 1 \\ -1 & 0 \\ 0 & -1\end{bmatrix} \lambda_{k}^{j} \Bigg|\Bigg|_{2} = 1
\end{align}
\end{subequations}
where $j = \{1,2,3...M\}, j \neq i$ and $k \in \{1,2,...N \}$. Hence for each agent $i$, there would be $M-1$ agents acting as obstacles which can be formulated as $M-1$ obstacle constraints similar to the above equation. We see that this formulation leads to a coupled constraint between neighbouring agents which is not separable. We intend to solve these problems with consensus based algorithm for distributed optimization.

\subsubsection{Distributed Optimal Control Problem}\label{local_prob_init}
Having formulated the global objective and constraints, we can now write down a distributed problem with each agent solving the following finite time optimal control problem

\begin{equation}\label{dist_prob_eq}
\begin{split}
    \min_{\textbf{s},\textbf{u},\lambda,\alpha} \sum_{k=0}^{k=N-1} &c(s(k),u(k)) + \kappa.\alpha_{k} + V(s(N)) \\
    s.t. \hspace{4pt} & s(k+1) = A_{k}.s(k) + B_{k}.u(k)\\
    \Bigg( &\begin{bmatrix}1 & 0\\ 0 & 1 \\ -1 & 0 \\ 0 & -1\end{bmatrix} (\Tilde{s}^{(i)}(k) - \Tilde{s}^{(j)}(k)) - \Delta \Bigg)^{T} \lambda_{k}^{j} \geq -\alpha_{k}^{j} \\
    & \Bigg|\Bigg|\begin{bmatrix}1 & 0\\ 0 & 1 \\ -1 & 0 \\ 0 & -1\end{bmatrix} \lambda_{k}^{j} \Bigg|\Bigg|_{2} = 1 \\
    &j \in  \{1,2,3...M \}, j \neq i, k \in \{1,2,3...N \}
\end{split}
\end{equation}

where the objective depends only on the state of the agent but there is a presence of a coupling constraint between neighbouring states.

\subsection{ADMM Based Distributed Optimization}

In this section we discuss the Alternating Direction Method of Multipliers (ADMM) for solving the Distributed MPC problem. The detailed proof and properties of the ADMM based Distributed MPC method can be found in \cite{Rostami2017ADMMbasedDM} 
\subsubsection{System Description and Notations}
We first discuss the system description along with the notations that we use for the method. The ADMM approach that we follow requires each agent of the network to store a local copy of the optimization variables associated with other agents in the network. Hence the optimization vector for agent $i$ is defined as
\begin{subequations}
\begin{align}
    & v^{i} = [\textbf{s}^{i}, \textbf{u}^{i}] = \big( [\textbf{s}_{j}^{i}, \textbf{u}_{j}^{i}] \big)_{j \in \mathcal{M}} \\
    &\textbf{s}_{j}^{i} = [s_{j}^{i}(0), s_{j}^{i}(1),s_{j}^{i}(2)...s_{j}^{i}(N)] \\
    &\textbf{u}_{j}^{i} = [u_{j}^{i}(0), u_{j}^{i}(1),u_{j}^{i}(2)...u_{j}^{i}(N-1)]
\end{align}
\end{subequations}
where $[\textbf{s}_{j}^{i}, \textbf{u}_{j}^{i}]$ denotes the local copy of the optimization variable (states and actions over the horizon) of agent $j$ for the agent $i$. $\mathcal{M}$ denotes the set containing the index of all agents in the network. Augmenting the local optimization variable with copies of the optimization variable of other agent allows us to completely decompose the global problem into sub-problems for each agent and hence solve the problem in a distributed fashion.   
\subsubsection{Consensus Constraint}
After having divided the problem into sup-problems, we need to ensure that the values of the local copies and the global optimization variables match each other. In order to achieve this, a consensus constraint is introduced for each of the agent.
\begin{equation}
    v^{i} - \Bar{v}^{i} = 0
\end{equation}
Here $\Bar{v}^{i}$ is defined as the network average variable which stores the average of all local copies of the optimization variables across different agents.
\begin{equation} \label{stacking}
    \Bar{v}^{i} = (\Bar{v}_{j})_{j \in \mathcal{M}}
\end{equation}
\begin{equation} \label{averaging}
    \Bar{v}_{i} = \frac{1}{M} \sum_{j \in \mathcal{M}} \Bar{v}_{i}^{j}
\end{equation}
The values of $\Bar{v}_{i}^{j}$ are obtained after the optimization procedure for agent $j$. Once the optimization procedure is completed for each of the agent, the resulting optimization variables are averaged to update the network average variable $\Bar{v}^{i}$ for each of the agent. The ADMM algorithm therefore iteratively alternates between local agent optimization and global averaging until consensus is achieved.
\subsubsection{Augmented Lagrangian Formulation}
For the ADMM algorithm, the local optimization problem as discussed in Section \ref{local_prob_init} has to be modified. We have to formulate a augmented lagrangian objective that takes into consideration the consensus constraint and penalizes values too far away from the network average. The augmented objective can now be written as
\begin{equation}
    \mathcal{L}(v^{i}, \Bar{v}^{i}, \gamma^{i}) = \sum_{k=0}^{k=N-1} C(v_{i}^{i}(k)) + \gamma^{iT}(v^{i}-\Bar{v}^{i}) + \frac{\rho}{2} ||(v^{i}-\Bar{v}^{i}||_{2}^{2}
\end{equation}
where $C(.)$ is as defined as in section \ref{single_obj}. $\gamma^{i}$ denotes the Lagrange multiplier for agent $i$ and $\rho$ denotes the parameter that penalizes deviation from network average.
Now at each iteration, we optimize the augmented lagrangian objective to obtain the local solution vector.
\begin{equation}\label{augmented_opt}
\begin{split}
    &v^{i+} = arg\min_{v^{i}} \mathcal{L}(v^{i}, \Bar{v}^{i}, \gamma^{i}) \\
    &v^{i} = [\textbf{s}^{i}, \textbf{u}^{i}] = \big( [\textbf{s}_{j}^{i}, \textbf{u}_{j}^{i}] \big)_{j \in \mathcal{M}} \\
    &s.t. \hspace{4pt}  s^{i}_{j}(k+1) = A_{k}.s_{i}^{i}(k) + B_{k}.u_{i}^{i}(k)\\
    & \Bigg( \begin{bmatrix}1 & 0\\ 0 & 1 \\ -1 & 0 \\ 0 & -1\end{bmatrix} (\Tilde{s}^{i}_{i}(k) - \Tilde{s}^{i}_{j}(k)) - \Delta \Bigg)^{T} \lambda_{k}^{i} \geq -\alpha_{k}^{i} \\
    & \Bigg|\Bigg|\begin{bmatrix}1 & 0\\ 0 & 1 \\ -1 & 0 \\ 0 & -1\end{bmatrix} \lambda_{k}^{i} \Bigg|\Bigg|_{2} = 1 \\
    &j \in  \{1,2,3...M \}, j \neq i, k \in \{1,2,3...N \}
\end{split}
\end{equation}
The optimal solution of the above problem is then transmitted to other agents in order to form the network average for the next iteration.
\subsubsection{ADMM Algorithm}
The overall ADMM algorithm can now be written including the multiplier update step as shown in Algorithm \ref{alg:admm_algo}.

\subsection{Pre-computing initial guess}
Due to the non-convexity of the problem, the computational load is relatively high for the solution. We try to mitigate this by using pre-computed initial guess solutions as guiding trajectories for the solver. At every timestep, we run a sampling based motion planning scheme based on Rapidly Exploring Random Trees (RRT) from the current state to the goal state and pass the first $N$ states as the initial guess to the solver. This accelerates the optimization process and also prevents locally minimum solutions in many cases.

\section{ALGORITHM}
\begin{algorithm}\label{admm_algo}
\caption{ADMM Algorithm}\label{alg:admm_algo}
\begin{algorithmic}[1]
\State For each agent i:
\State $\gamma^{i} = 0, \Bar{v}^{i} = 0$
\While{Not Converged}
\State calculate $v^{i+}$ from \eqref{augmented_opt}
\State Average all local copies from \eqref{averaging}  
\State Update network average variable $\Bar{v}_{i}^{+}$ using \eqref{stacking}
\State $\gamma^{i+} = \gamma^{i} + \rho (v^{i+} - \Bar{v}^{i+})$
\EndWhile
\end{algorithmic}
\end{algorithm}
\begin{algorithm}[!tbh]
\caption{DMPC Algorithm}\label{alg:dmpc_algo}
\begin{algorithmic}[1]
\State Initialize agents with current state and goal state
\State Initialize optimization parameters $\rho, N, Q$
\While{Goal not reached}
\Procedure{RRT}{$s_{0}, s_{g}$} \Comment{compute initial guess}\EndProcedure
\State For each agent $i$:
\State Solve for optimal trajectory using Algorithm \ref{alg:admm_algo} with initial guess
\State Update current state $s_{0}$ using consensus solution
\EndWhile
\end{algorithmic}
\end{algorithm}

\section{EXPERIMENTS AND RESULTS}\label{exp_and_res}
\begin{figure*}
    \centering
    \begin{subfigure}[b]{0.45 \textwidth}
    \centering
    \includegraphics[width=\textwidth]{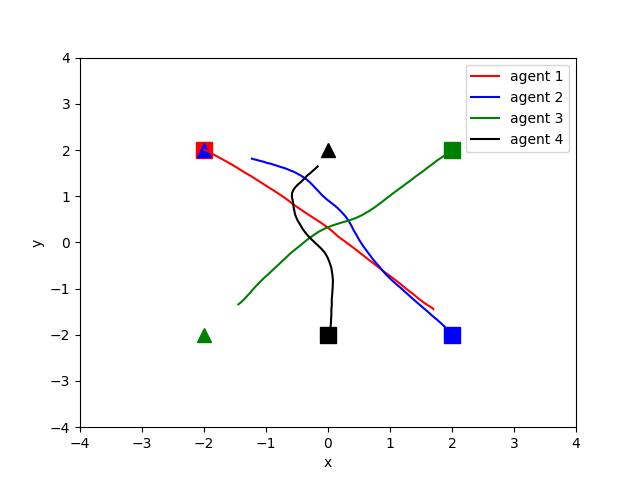}
    \caption{2D Environment}
    \label{traj_2d}
    \end{subfigure}
    \begin{subfigure}[b]{0.45 \textwidth}
    \centering
    \includegraphics[width=\textwidth]{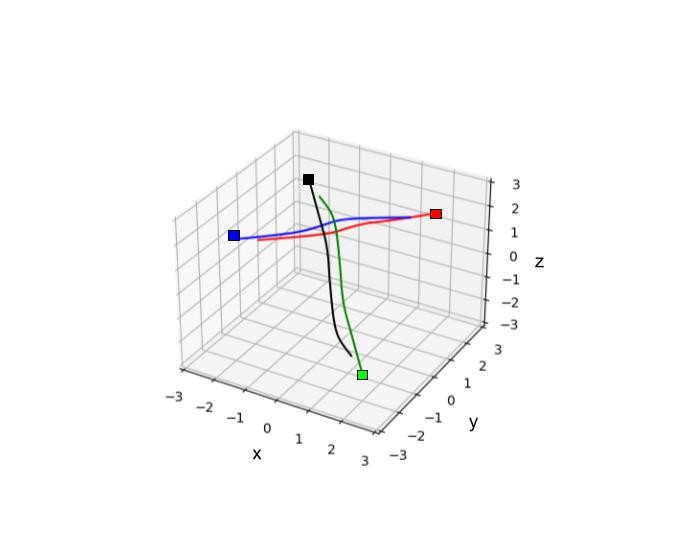}
    \caption{3D Environment}
    \label{3d}
    \end{subfigure}
    \caption{Simulation Results}
    \label{sim_result}
\end{figure*}
\subsection{Parameters}
For the simulation study, we consider $M = 4$ agents to validate the algorithm. For the distributed MPC problem, we consider $N = 10$ as the horizon length with a discretization time of $\Delta T = 0.1s$. We vary the values of $\Delta$ within a set of admissible values $\Delta \in \{0.1, 0.3, 0.5\}$. The cost matrices as discussed in section \ref{single_obj} are defined as $Q = 0.1*I_{2n}$, $R = I_{n}$ and $Q_{f} = I_{2n}$ $\forall n \in \{2,3\}$ denoting the 2D and 3D system respectively. 

\subsection{Simulation results}

\begin{figure}[!tbh]
    \centering
    \includegraphics[width=0.47 \textwidth]{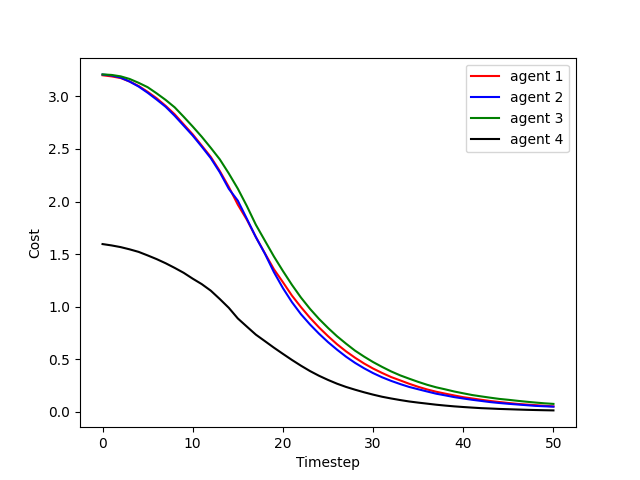}
    \caption{Convergence of algorithm}
    \label{convergence}
\end{figure}

We are able to validate the algorithm on a simulation environment and the results shown in Fig. \ref{sim_result} suggest the success of the algorithm in achieving collision avoidance and goal reaching for each of the agent. 

\subsection{Variation with $\Delta$}

\begin{figure}[!tbh]
    \centering
    \includegraphics[width=0.47 \textwidth]{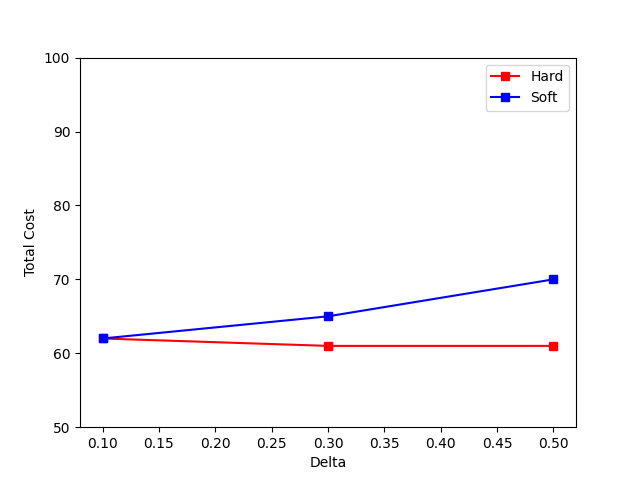}
    \caption{Variation of performance with $\Delta$}
    \label{delta_var}
\end{figure}

As observed in Fig. \ref{delta_var}, the performance slightly deteriorates with an increase in the value of the parameter $\Delta$ or the size of the cube around each agent. This is due to the conservative solutions associated with high value of the parameter. The effect of this parameter is more significant in dense environments leading to in-feasibility issues with a hard collision constraint formulation.

\subsection{Hard vs Soft constraints}
We observe that soft constraints provide slightly sub-optimal solution as compared to the hard constraints as seen in Fig. \ref{delta_var}. This is because it finds a sub-optimal solution which may not be feasible in the hard constraint formulation. Soft constraints are essential in dense environments as a conservative modelling of obstacles in tight spaces will lead to frequent infeasible solutions. Soft constraint formulation leads to elevated computational time as compared to hard formulation because of the non-convex equality constraint.

\subsection{Limitations and possible solutions}
The current method has a relatively high computational load because of the collision avoidance constraint being solved at each time step. Incorporating an on-demand collision avoidance strategy as discussed in \cite{Luis2020OnlineTG} can help to mitigate this for high frequency real time operations.

\section{CONCLUSIONS AND FUTURE WORK}\label{conc_and_future}
In this work, we present an extension to the work on optimization based collision avoidance for multi-agent point to point transition tasks. We model the agents as convex polyhedron obstacles and formulate the collision avoidance constraint using a reformulation method discussed in optimization based collision avoidance literature. Using the separable nature of the global objective, we are able to formulate the problem in a distributed manner with separable objective but coupled constraints among neighbouring agents. We solve the given problem using a consensus based ADMM algorithm.

We observe that the computational loads without code optimization is relatively high and hence we experiment with methods like online pre-computation of initial guess for the solver. Such modifications provide several advantages like lower computational times, high reliability and minimal deviation from optimal solution. We also observe the effects of parameter changes and changes in the reformulation methods on the nature of solution in terms of conservativeness, feasibility and optimality. All these ablations help us conclude on the usefulness of the method and the possibility of real-time application of such distributed methods for multi-agent robotic systems.

One of the key challenges not addressed by this work is handling of uncertainty within the system. For multi-agent systems and especially for aerial swarms, there are a lot of uncertainties associated with the system. When different UAVs come close together, the mutual interaction between them induces certain uncertainty in the dynamic model prediction. UAVs flying close to the ground also face uncertainty due to the downwash effect. Various problems involving aerial swarms also include suspended payloads to the UAVs which create more deviation from model predictions. All such uncertainties can be modelled as both random and systemic uncertainties. As a future work, we aim to model these uncertainties within the current formulation and develop Distributed Stochastic MPC based methods to solve the optimization problem robustly.

\addtolength{\textheight}{-12cm}   







\bibliographystyle{IEEEtran}
\bibliography{citations}


\end{document}